%% file: main.tex
\pdfoutput=1
\def\year{2022}\relax
\documentclass[letterpaper]{article} 
\input{packages}

\title{Cross-Domain Empirical Risk Minimization for \\ Unbiased Long-tailed Classification}

\author{%
 \textbf{Beier Zhu}\textsuperscript{1} \quad \textbf{Yulei Niu}\textsuperscript{1}\thanks{Corresponding author} \quad \textbf{Xian-Sheng Hua}\textsuperscript{2} \quad \textbf{Hanwang Zhang}\textsuperscript{1} \\
\small \textsuperscript{1}Nanyang Technological University\quad \quad  \textsuperscript{2}Damo Academy, Alibaba Group\\
\tt\small beier002@e.ntu.edu.sg, yn.yuleiniu@gmail.com\\ \tt\small  xiansheng.hxs@alibaba-inc.com \quad hanwangzhang@ntu.edu.sg
 }

\begin{document}

\maketitle

\input{sections/0-abs}
\input{sections/1-intro}
\input{sections/2-related}
\input{sections/3-algorithm}

\input{sections/4-justification}
\input{sections/5-experiments}
\input{sections/6-conclusion}
\input{sections/7-ack}

\input{sections/a1-proof}
\input{sections/a2-implementation}
\input{sections/a3-experiments}
\bibliography{aaai22}

\end{document}

%% file: packages.tex
\usepackage{aaai22}  
\usepackage{times}  
\usepackage{helvet}  
\usepackage{courier}  
\usepackage[hyphens]{url}  
\usepackage{graphicx} 
\usepackage{natbib}  
\usepackage{caption} 
\DeclareCaptionStyle{ruled}{labelfont=normalfont,labelsep=colon,strut=off} 
\usepackage{algorithm}
\usepackage{algorithmic}

\usepackage{newfloat}
\usepackage{listings}
\floatstyle{ruled}
\newfloat{listing}{tb}{lst}{}
\floatname{listing}{Listing}

\usepackage[utf8]{inputenc} 
\usepackage[T1]{fontenc}    
\usepackage{hyperref}       
\usepackage{url}            
\usepackage{booktabs}       
\usepackage{amsfonts}       
\usepackage{nicefrac}       
\usepackage{microtype}      
\usepackage{xcolor}         
\usepackage{wrapfig}
\usepackage{multirow}
\usepackage{enumitem}
\usepackage{bm}
\usepackage{amsmath}
\usepackage{caption}
\usepackage{nccmath} 
\usepackage{bibentry}
\usepackage{adjustbox}

\newcommand{\ie}{\textit{i.e.}}
\newcommand{\eg}{\textit{e.g.}}

\newcommand{\cf}{\textit{cf. }}

\newcommand\tbf{\textbf}
\newcommand{\niu}[1]{\textcolor{red}{#1}}

\newcommand{\F}{\text{imba}}
\newcommand{\CF}{\text{ba}}
\newcommand{\pf}{p^\F}
\newcommand{\pcf}{p^\CF}
\newcommand{\wf}{w^\F}
\newcommand{\wcf}{w^\CF}

\newcommand{\CE}{X\!E}
\newcommand{\CEF}{\CE^\F}
\newcommand{\CECF}{\CE^\CF}

\newcommand{\pX}{P(X)}
\newcommand{\px}{P(x)}
\newcommand{\pXx}{P(X\!=\!x)}

\newcommand{\pygdox}{P(y|do(x))}

\newcommand{\pYgdoXx}{P(Y|do(X=x))}
\newcommand{\pYygdoXx}{P(Y\!=\!y|do(X\!=\!x))}
\newcommand{\ygs}{y_{s=1}}
\newcommand{\ygns}{y_{s=0}}

\newcommand{\Loss}{\mathcal{L}}
\newcommand{\Lossxy}{\Loss(y,f(x))}
\newcommand{\Lossxys}{\Loss(\ygs,f(x))}
\newcommand{\Lossxyns}{\Loss(\ygns,f(x))}
\newcommand{\Lossxyss}{\Loss(y_{s},f(x))}
\newcommand{\pxys}{P(x,y,S=1)}
\newcommand{\pxyns}{P(x,y,S=0)}
\newcommand{\pxyss}{P(x,y,s)}
\newcommand{\pxgs}{P(x|S=1)}
\newcommand{\pxgns}{P(x|S=0)}
\newcommand{\pxgss}{P(x|S=s)}
\newcommand{\pygdoxs}{P(y|do(x),S=1)}
\newcommand{\pygdoxns}{P(y|do(x),S=0)}
\newcommand{\pdoxys}{P(do(x),y,S=1)}
\newcommand{\pdoxyns}{P(do(x),y,S=0)}
\newcommand{\pdoxs}{P(do(x),S=1)}
\newcommand{\pdoxns}{P(do(x),S=0)}
\newcommand{\ps}{P(S=1)}
\newcommand{\pns}{P(S=0)}
\newcommand{\pdoxgs}{P(do(x)|S=1)}
\newcommand{\pdoxgns}{P(do(x)|S=0)}
\newcommand{\psgx}{P(S=1|do(x))}
\newcommand{\pnsgx}{P(S=0|do(x))}
\newcommand{\la}{\!\leftarrow\!}

\newcommand{\pxs}{P(x,S=1)}
\newcommand{\pxns}{P(x,S=0)}

\usepackage{pifont} 

\newcommand{\cmark}{\ding{51}}%

\newlength\savewidth\newcommand\shline{\noalign{\global\savewidth\arrayrulewidth
  \global\arrayrulewidth 1pt}\hline\noalign{\global\arrayrulewidth\savewidth}}
  

%% file: sections/0-abs.tex
\begin{abstract}
We address the overlooked unbiasedness in existing long-tailed classification methods: we find that their overall improvement is mostly attributed to the biased preference of ``tail'' over ``head'', as the test distribution is assumed to be balanced; however, when the test is as imbalanced as the long-tailed training data---let the test respect Zipf's law of nature---the ``tail'' bias is no longer beneficial overall because it hurts the ``head'' majorities. In this paper, we propose Cross-Domain Empirical Risk Minimization (xERM) for training an unbiased  model to achieve strong performances on both test distributions, which empirically demonstrates that xERM fundamentally improves the classification by learning better feature representation rather than the ``head vs. tail'' game. 
Based on causality, we further theoretically explain why xERM achieves unbiasedness: the bias caused by the domain selection is removed by adjusting the empirical risks on the imbalanced domain and the balanced but unseen domain. Codes are available at {\small \url{https://github.com/BeierZhu/xERM}}.

\end{abstract}

%% file: sections/1-intro.tex
\section{Introduction}\label{sec:intro}
When the training data distribution is long-tailed, \eg, thousands of samples per ``head'' class vs. just a few samples per ``tail'' class, the resultant classification model will inevitably downplay the minor ``tail'' and overplay the major ``head''. Therefore, to lift up the ``tail'' performance, recent methods focus on the trade-off between the ``head'' and ``tail'' contributions, \eg, sample re-weighting~\citep{cui2019class,lin2017focal}, balanced training loss~\citep{Decouple, DisAlign}, and ``head'' causal effect removal~\citep{TDE}. As you may expect, the ``tail'' rise costs the ``head'' fall; but fortunately, the overall gain is still positive. The reasons are two-fold. First, thanks to the abundant ``head'' training samples, the strong ``head'' features 
can afford the trade-off, making the ``head'' performance drop much less significant than the ``tail'' rise. Second, the test distribution is often assumed to be balanced, \ie, ``head'' and ``tail'' have the same number of samples, so the ``head'' fall is not amplified by the balanced sample number.

We conjecture that the above ``overall improvement'' on the balanced test may cover up the fact that today's long-tailed classification methods are still \emph{biased} and far from truly ``improved''. Figure~\ref{fig:fi1} (a.1) and (b.1) show an evidence. Regardless of the test distributions, balanced or imbalanced, an imbalanced model (vanilla cross-entropy trained, XE, orange line) always predicts the class distribution shaped as its own model prior, which is long-tailed. As a comparison, a balanced model (a SOTA method TDE~\citep{TDE}, blue line) always predicts the distribution to cater the tail. This observation implies that both of them are biased: the imbalanced model is ``head'' biased; the balanced one is ``tail'' biased, which merely replaces the ``head'' bias with the ``tail'' bias.
Therefore, we expect an ``unbiased'' model's prediction to  treat each class fairly and respects both test distributions, \ie, the prediction distribution follows the shape of the ground-truth one (\eg, green line).
Figure~\ref{fig:fi1} (a.2) and (b.2) provide another evidence based on the true positive (TP) and the false positive (FP) statistics. We observe that the balanced model TDE tends to predict more samples as tail classes. On the contrary, the imbalanced model XE prefers to predict more samples as head classes. The  preference on head or tail classes comes with a large number of FP samples, which lowers the precision score. In contrast, our xERM has no significant preference between head and tail classes, which achieves high precision under both test distributions.
Only through the above unbiasedness, we can diagnose whether a model truly improves long-tailed classification,
but not by playing a ``prior flipping'' game (\cf Section~\ref{sec:feature}).

\input{images/fig1}

In fact, it is challenging to achieve unbiasedness across the two completely different test distributions due to the domain selection issue~\citep{pearl2011transportability}: the two distributions would never co-exist in training --- once one is observed, the other is gone --- let alone adjusting a fair prior for both of them. In this paper, we propose a novel training paradigm called Cross-Domain Empirical Risk Minimization (xERM) to tackle this challenge. xERM consists of two ERM terms, which are the cross-entropy losses with different supervisions: one is the ground-truth label from the seen imbalanced  domain, and the other is the prediction of a balanced model.
The imbalanced domain empirical risk encourages the model to learn from the ground-truth annotations from the imbalanced distribution, which favors the head classes. The balanced domain empirical risk encourages the model to learn from the prediction of a balanced model, which imagines a balanced distribution and prefers the tail classes. These two risks are weighted to take advantage of both, \ie, protect the model training from being neither too ``head biased'' nor too ``tail biased'', and can achieve the unbiasedness (\cf Figure~\ref{fig:fi1}).
Furthermore, we propose a causal theory to ground xERM, which removes the confounding bias caused by domain selection (\cf Section~\ref{sec:just}).

We conduct experiments on several long-tail classification benchmark datasets: CIFAR100-LT~\cite{cui2019class}, Places365-LT~\cite{OLTR}, ImageNet-LT~\cite{OLTR}, and iNaturalist 2018~\cite{inaturalist}. Experimental results show that xERM outperforms previous state-of-the-arts on both long-tailed and balanced test sets, which demonstrates that the performance gain is not from catering to the tail. Further qualitative studies show that the xERM helps with better feature representation.

We summarize our contributions as follows:
\vspace{-1mm}
\begin{itemize}
\item We point out the overlooked \emph{unbiasedness} in long-tailed classification: models should perform well on both imbalanced and balanced test distributions. Compared to the conventional evaluations on the balanced test only, the unbiasedness thoroughly diagnoses whether a model is spuriously improved by simply flipping the bias.

\item We propose xERM that achieves the unbiasedness. On various long-tailed visual recognition benchmarks, we show that xERM constantly achieves strong performances across different test distributions.

\item We provide a theoretical analysis for xERM which reduces the confounding bias based on the causal theory.
\end{itemize}

%% file: images/fig1.tex
\begin{figure*}
    \centering
    \includegraphics[width=0.95\textwidth]{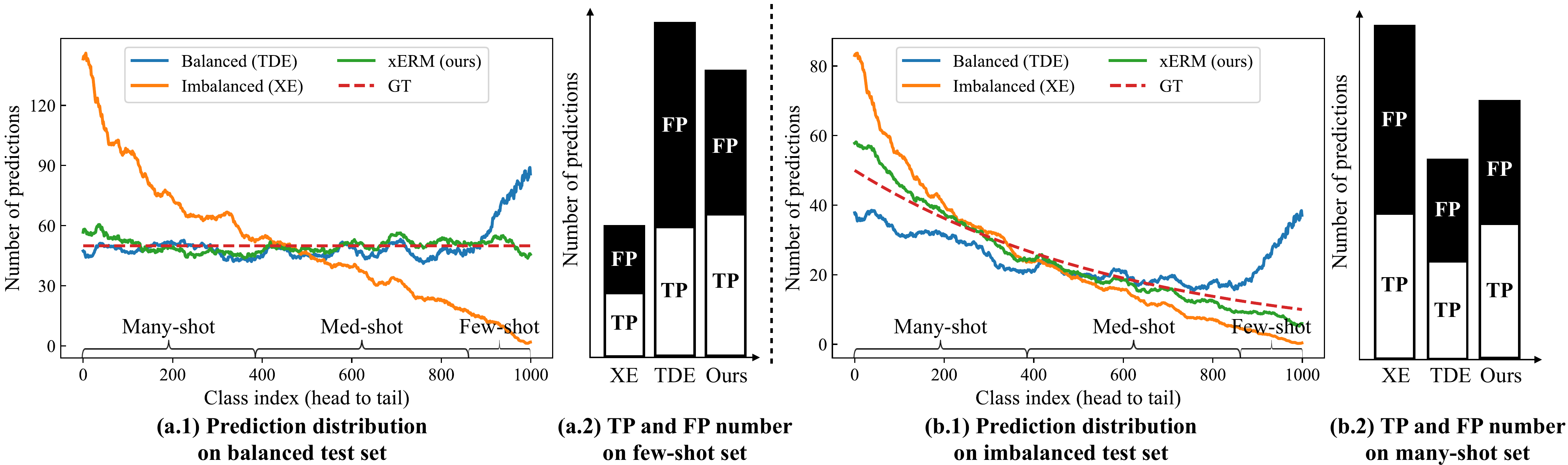}
    \caption{Prediction distributions of imbalanced model (XE), balanced model (TDE, \cite{TDE}) and our xERM model on (a.1) balanced test and (b.1) imbalanced test (long-tailed) of ImageNet-LT~\citep{OLTR}. Interestingly, the abrupt tail rise of the blue lines shows an obvious ``tail" preference. The true positive (TP) and false positive (FP) number of the predictions on (a.2) few-shot subset of the balanced test set and  (b.2) many-shot subset of the imbalanced test set.}
    \label{fig:fi1}
\end{figure*}

%% file: sections/2-related.tex
\section{Related Work}

\textbf{Long-tailed classification}.
Early works focus on re-balancing the contribution of each class in the training phase, which can be divided into two strategies: re-sampling the data to balance the class frequency~\cite{buda2018systematic,drumnond2003class,shen2016relay, sarafianos2018deep,japkowicz2002class}, and re-weighting the loss of classes~\cite{tan2020equalization,cui2019class,lin2017focal,khan2017cost} to increase the importance of tail classes. Nevertheless, both strategies suffer from under-fitting/over-fitting problem to head/tail classes. Recent studies shows a trend of decoupling long-tailed classification into two separate stages: representation learning and classifier learning. Decouple~\cite{Decouple} learns high-quality representation with natural (instance-balanced) sampling, and achieve strong classification performance by adjusting classifier with balanced learning. \citet{BBN} arrives at a similar conclusion by proposing a bilateral-branch network, where one branch is trained with natural sampling and the other uses balanced sampling. The decoupled learning idea is also adopted in~\cite{logitsAdjust, LADE, TDE, DisAlign}, where different classifier adjustment methods are proposed. There is a line of post-hoc logits adjustment by subtracting the training prior from the predicted logits~\cite{logitsAdjust, LADE}. \citet{TDE} removes the ``head'' causal effect to get the adjusted logits. DisAlign~\cite{DisAlign} modifies the original logits by adding an extra learnable layer to the output layer. 
Apart from decoupled learning, \citet{liu2020deep} addresses the long-tailed problem by transferring head distributions to tail; distilling balanced student models from imbalanced trained teacher models~\cite{DiVE, BKD}. Although all the above methods manifest an overall accuracy improvement, as we discussed in Section~\ref{sec:intro}, they are indeed biased.

\noindent \textbf{Causal Inference.} Causal inference~\citep{pearl2016causal, peters2017elements} has been studied in a wide spectrum, such as economics, politics and epidemiology~\citep{heckman1979sample, keele2015statistics, robins2001data}.
Recently, causal inference has also shown promising results in various computer vision tasks, \eg, pursuing pure direct effect in VQA~\citep{niu2021counterfactual}, back-door and front-door adjustment for captioning~\citep{yang2020deconfounded} and counterfactual inference for zero-shot recognition~\citep{yue2021counterfactual}. Our algorithm can be viewed as a practical implementation and complement of the recent empirical risk minimization of learning causal effect~\citep{jung2020learning} for confounding effect removal.

\noindent\textbf{Bias mitigation.} A bias mitigation work related to us is LfF~\cite{nam2020learning}. LfF used a reweighting-based method to deal with bias in out-of-distribution robustness, and used the cross-entropy loss to calculate the weights. The main differences are as follows. First, LfF focused on biased dataset with spurious correlations, while we focused on long-tailed task with label prior shift. 
Second, LfF weights the losses at the sample-level, while xERM weights the losses at the supervision-level. Third, LfF calculates the weights using a biased model and a target debiased model, while xERM calculates the weights using fixed balanced and imbalanced models which can be merged into one.

%% file: sections/3-algorithm.tex
\section{The Algorithm: xERM}\label{sec:alg}
\noindent\textbf{Input:} Long-tailed training samples denoted as the the pairs $\{(x,y)\}$ of
an image $x$ and its one-hot label $y$.

\noindent\textbf{Output:} Unbiased classification model $f$.
\begin{enumerate}[label=\textbf{Step \arabic*:}, wide, labelwidth=!, labelindent=0pt]
\item \textbf{Learn an imbalanced and a balanced model.} We learn a conventional model $\pf(y|x)$ on the imbalanced training data as the imbalanced model, which favors the ``head'' bias. Then, we learn a balanced model $\pcf(y|x)$, which favors the ``tail'' bias. The implementations of the two models are open. See Section~\ref{sec:exp} and Appendix for our choices.

\item \textbf{Estimate the adjustment weights.}
\begin{enumerate}[label=, leftmargin=0pt]
\item
Imbalanced Domain ER weight:
\begin{equation}\label{eq:wf}
\wf=\frac{(\CEF)^\gamma}{(\CEF)^\gamma+(\CECF)^\gamma},
\end{equation}

\item 

Balanced Domain ER weight:
\begin{equation}\label{eq:wcf}
\wcf=1-\wf = \frac{(\CECF)^\gamma}{(\CEF)^\gamma+(\CECF)^\gamma},
\end{equation}
where 
$\CEF = -\sum_i y_i\log \pf (y_i|x)$ is 
the cross-entropy of the imbalanced model's predictions
(similar for $\CECF$), the subscript $i$ denotes the $i$-th class label, and $\gamma>0$ is a scaling hyper-parameter. 

\end{enumerate}

\item \textbf{Minimize the cross-domain empirical risk}
\begin{enumerate}[label=, leftmargin=0pt]
\item
\begin{flalign}\label{eq:fer}
\text{\small Imbalanced Domain ER:}&\ \mathcal{R}^{\F}(f)=-\wf\sum_i y_i\log f_i(x),&
\end{flalign}
where $y_i$ and $f_i$ are the ground-truth and the predicted label for $i$-th class, respectively.

\item
\begin{flalign}\label{eq:cfer}
\text{\small Balanced Domain ER:}&\qquad\mathcal{R}^{\CF}(f)=-\wcf\sum_i \hat{y}_i\log f_i(x),&
\end{flalign}

where $\hat{y_i}=\pcf(y_i|x)$ denotes the balanced prediction for $i$-th class. 
The overall empirical risk minimization:
\begin{equation}
    \mathcal{R}(f)=\mathcal{R}^{\F}(f)+\mathcal{R}^{\CF}(f).
\end{equation}
\end{enumerate}

\end{enumerate}

%% file: sections/4-justification.tex
\section{Justification}\label{sec:just}

\textbf{Causal Graph.} We first construct the causal graph for long-tailed classification to analyze the causal relations between variables. As shown in Figure~\ref{fig:causal-graph} (a), the causal graph contains three nodes: $X$: image, $Y$: prediction, $S$: selection variable. Each value of $S$ corresponds to a domain, and switching between the domains will be represented by conditioning on different values of the $S$ variable. In long-tailed classification, let $S=0$ denote the balanced domain and $S=1$ denote the imbalanced (long-tailed) domain.\footnote{Technically, the possible values of $S$ are infinite and far more than only confined to 0 (balance) or 1 (long-tail). For backdoor adjustment, the summation should have considered every possible domain of $S$. 
In this paper, we focus on the long-tailed and balanced domains because these two cases are the most representative and general scenarios for classification. A fine-grained definition of $S$ leads to heavy computation, which is impractical for training.}
\input{images/causal-graph}

\noindent $X \rightarrow Y$. This path indicates that the model predicts the label based on the image content. 

\noindent $S \rightarrow X$. This path indicates that an image is sampled according to the selected domain, \eg, imbalanced domain is prone to include images from head classes.

\noindent $S \rightarrow Y$. This path indicates that the predicted label distributions follow their own training domain prior. For example, in Figure \ref{fig:fi1}, an imbalanced model XE always predicts a long-tailed label distribution regardless of the test distributions, and the balanced model TDE shows an obvious ``tail'' preference across different test distributions.

Note that the back-door path $X \la S \rightarrow Y$ contributes a spurious correlation between $X$ and $Y$~\citep{pearl2016causal}, where $S$ acts as a confounder. The existence of back-door path is the key challenge for achieving unbiasedness across balanced and imbalanced test data. Simply learning by $P(Y|X=x)$ will inevitably include such spurious correlation, which is not what we seek to estimate, and lead to a biased model. Therefore, we instead estimate $P(Y|do(X=x))$ which can eliminate the confounding bias to obtain an unbiased model. 


\noindent \textbf{Empirical Risk on Intervened Distribution.} The empirical risk minimization for learning causal effect $P(Y|do(X=x))$ has been well formulated in~\cite{jung2020learning}. We define the empirical risk $\mathcal{R}$ of the estimator $f$ on the intervened distributions  as: 

\begin{equation}\label{eq:rf}
\begin{aligned}
    \mathcal{R}(f)&=\mathbb{E}_{x\sim\pX,y\sim\pYgdoXx}\Lossxy \\
                  &=\sum_x\sum_y\Lossxy\pygdox\px,
\end{aligned}
\end{equation}
where $\Lossxy$ is formulated as the standard cross-entropy. By $do$-operation (causal intervention), we cut off the in-paths to $X$ and transform the causal graph in Figure~\ref{fig:causal-graph} (a) to Figure~\ref{fig:causal-graph} (b). 
Different from traditional empirical risk that samples $(x,y)$ from $P(X,Y)$, we sample $(x,y)$ from the intervened distribution $P(do(X),Y)$. According to the rules of do-calculus~\cite{pearl2000causality}, we have $P(do(X=x))=P(X=x)$. Therefore, we can implement the sampling process drawn from the intervened distribution as two steps.
First, we sample $x$ from $\pX$. Second, we sample $y$ from the intervened conditional probability $\pYgdoXx$. In short, we denote $\pXx$ as $\px$ and $\pYygdoXx$ as $\pygdox$.  

The backdoor adjustment~\citep{pearl2000causality,pearl2016causal} allows us to remove $do(x)$ in $P(y|do(x))$ for easier estimation. We rewrite $P(y|do(x))$ \textit{w.r.t.} the confounder $S$ as: 

\begin{equation}\label{eq:pygdox}
\begin{aligned}
    \pygdox&=\sum\limits_{S=s\in\{0,1\}}P(y|x,S=s)P(S=s)  \\
           &=\frac{\pxys}{\pxgs}+\frac{\pxyns}{\pxgns}.
\end{aligned}
\end{equation}
A more detailed explanation of Eq. \eqref{eq:pygdox} is provided in Appendix. Here $S=1$ denotes the seen imbalanced domain, \ie, training samples, and $S=0$ denotes the unobserved balanced domain. Combining Eq.~\eqref{eq:pygdox} and Eq.~\eqref{eq:rf}, we have:

\begin{align}
\begin{split}
    \mathcal{R}(f)&=\sum_{(x,y)}\sum_{s\in\{0,1\}}\Lossxyss\frac{\px}{\pxgss}\pxyss\\
    &=\frac{1}{N}\sum_{(x,y)}[\underbrace{\Lossxys\frac{\px}{\pxgs}}_{\text{Imbalanced Domain ER}}\\
    &\quad\quad\quad\quad+\underbrace{\Lossxyns\frac{\px}{\pxgns}}_{\text{Balanced Domain ER}}],
\end{split}
\end{align}

\noindent {where $y_{s=1}$ denotes the ground-truth label from the seen imbalanced domain and $y_{s=0}$ denotes the supervision from the unseen balanced domain}. Here remain two questions. First, how to obtain $y_{s=0}$. As mentioned above, we cannot directly draw samples $(x,y_{s=0})$ from the unseen domain $S\!=\!0$. Fortunately, thanks to the causality-based framework~\citep{TDE}, this can be simulated by 
\textit{counterfactual inference} that deducts the ``head'' prior effect from the factual effect. In this way, we can obtain the counterfactual $(x,y_{s=0})$ based on the factual $(x,y_{s=1})$.
In particular, 
$y_{s=0}$ is estimated from the balanced classifier $\pcf(y|x)$ described in Step 1 and denoted as $\hat{y}$ in Step 3 of the algorithm. 
Second, how to obtain the sample weights $\frac{P(x)}{P(x|S)}$. 
Since $\frac{P(x)}{P(x|s)}=\frac{P(s)}{P(s|x)}$ and we assume $P(S\!=\!1)=P(S\!=\!0)$. Therefore, we have $\frac{P(x)}{P(x|s)}\propto\frac{1}{P(s|x)}$. 
In essence, they are \emph{inverse} probabilities of how likely the sample $x$ is from the imbalanced training domain ($S=1$) and from the balanced domain ($S=0$). Inspired by this, we propose to use the 
cross-entropy loss $\CE$ to delineate the discrepancy between the imbalanced and balanced domains and estimate the weights:
\begin{equation}\label{eq:wfcf}
\begin{split}
    \wf &\propto \frac{1}{P(S=1|x)}\propto{(\CEF)^\gamma},\\
    \wcf &\propto \frac{1}{P(S=0|x)}\propto{(\CECF)^\gamma}.
\end{split}
\end{equation}
Intuitively, the above equation says that if $x$ does not fit to the imbalanced model $\pf(y|x)$ (large $\CEF$) 
, it is not likely from the imbalanced training domain (small $P(S=1|x)$), and vice versa. 
By normalizing the weights to sum 1, we can easily derive Eq. \eqref{eq:wf} and \eqref{eq:wcf} from Eq. \eqref{eq:wfcf}.
A large hyper-parameter $\gamma\!>\!0$ would push $w$ towards 0 or 1 and we studied the effect of $\gamma$ in Section~\ref{sec:exp-ab}.


%% file: images/causal-graph.tex
\begin{figure}[h]
\centering
\vspace{-2mm}
\includegraphics[width=0.28\textwidth]{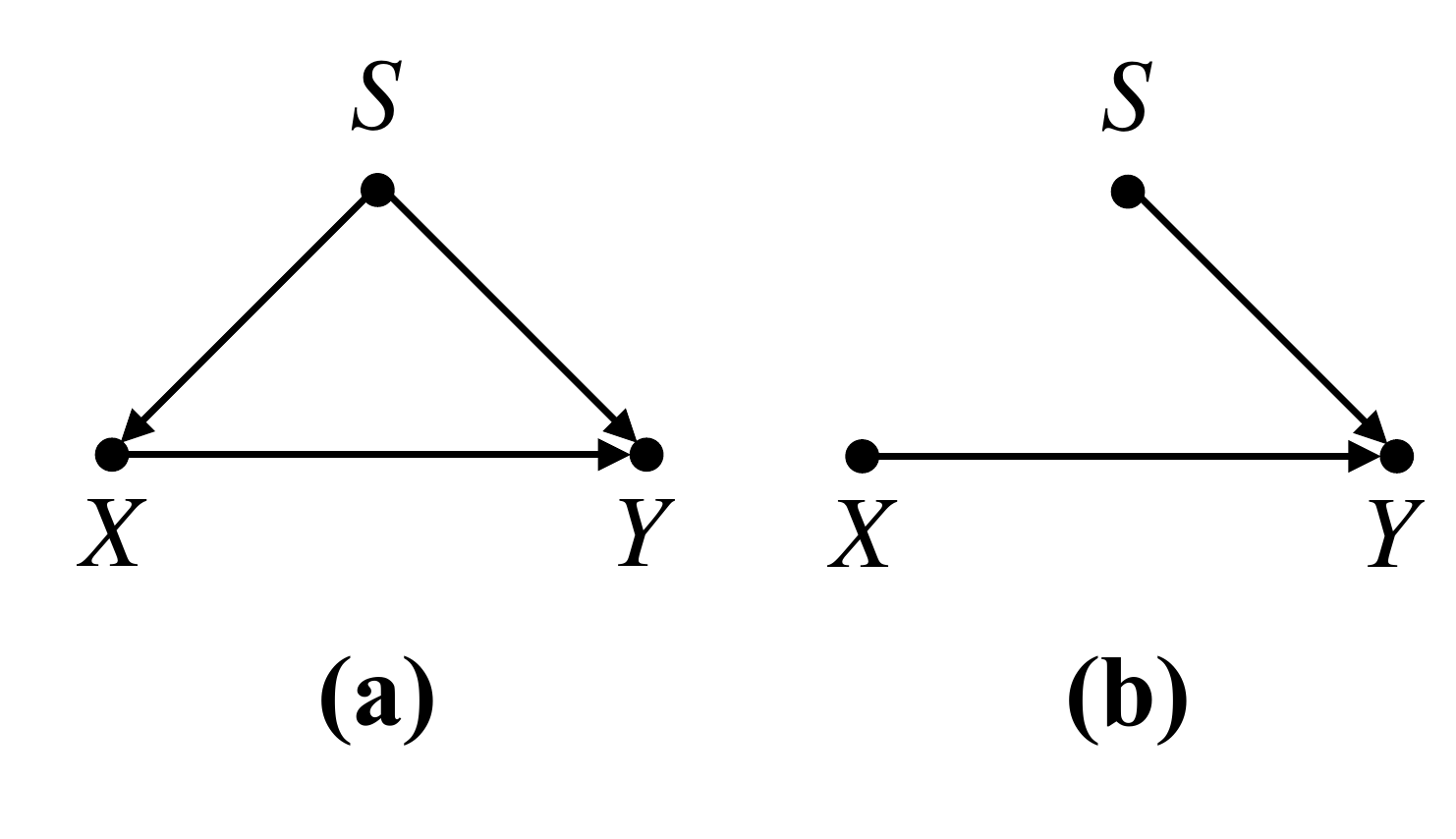}
\vspace{-4mm}
\caption{(a) The causal graph for long-tailed classification. The spurious correlation is induced as a backdoor $X \la S \rightarrow Y$ via the selection variable $S$. (b) The post-interventional model cuts off the arrow towards $X$ by $do(X)$.}
\label{fig:causal-graph}
\vspace{-2mm}
\end{figure}

%% file: sections/5-experiments.tex
\section{Experiments}\label{sec:exp}

\subsection{Setup}
\textbf{Datasets.} We conducted experiments on four long-tailed classification datasets: 
CIFAR100-LT~\cite{cui2019class}, Places365-LT~\cite{OLTR}, ImageNet-LT~\cite{OLTR}, and iNaturalist 2018~\cite{inaturalist}. 
The imbalance ratio (IB) is defined as $N_{\max}/N_{\min}$, where $N_{\max}$ ($N_{\min}$) denotes the largest (smallest) number of class samples. A larger IB denotes a more imbalanced split. Details of the datasets are in Table~\ref{tab:dataset}.
See Appendix for further details.
\input{tables/dataset}

\noindent\textbf{Backbones.} For fair comparisons, we used ResNet-32~\cite{he2016deep} as the backbone for CIFAR100-LT, ResNet-\{50, 152\} for Places365-LT, ResNeXt-50 for ImageNet-LT and ResNet-50 for iNaturalist18. For Places365-LT, we pretrained the backbones on ImageNet dataset as in~\citet{OLTR}. 

\noindent\textbf{Comparison methods.} We compare the proposed xERM models with state-of-the-arts such as BBN~\citep{BBN}, LDAM~\citep{LDAM}, LFME~\citep{MultiExpert}, BKD~\citep{BKD}, cRT, $\tau$-Norm, LWS~\citep{Decouple}, LADE, PC~\citep{LADE}, LA~\citep{logitsAdjust},TDE~\citep{TDE}, DiVE~\citep{DiVE} and DisAlign~\citep{DisAlign}.

The above mentioned methods use single models. Note that there is also a line of using ensemble models for long-tailed classification, \eg, RIDE~\cite{wang2020long} and CBD~\cite{CBD}. For fair comparisons, we will not include their results in the experiments. But, in Appendix, we show that our xERM framework can be easily applied to RIDE for validating its effectiveness on ensemble models. 

\noindent\textbf{Implementation.} We chose TDE and PC as baselines to implement xERM for two reasons. First, as described in Step 1 in Section 3, xERM requires an imbalanced model and a balanced model. Since TDE and PC are post-hoc logits adjustment methods, we can also obtain the imbalanced model effortlessly by disabling the post-hoc operation, which reduces the computation cost. Otherwise, we have to separately pre-train two models for the purposes. The second is performance. Balanced model with higher performance leads to a better xERM model. TDE and PC achieve stronger performance comparing to previous state-of-the-arts.
TDE first projects the feature from the penultimate layer to the orthogonal direction of the mean feature, then predicts the logits based on the projected feature. PC adjusts logits by subtracting the training prior from the predicted logits. We set the scaling parameter $\gamma$ in Eq.~\eqref{eq:wfcf} to 2 for CIFAR100-LT, 5 for Places365-LT and 1.5 for other datasets. Analysis of $\gamma$ effect is detailed in Section~\ref{sec:exp-ab}. See Appendix for more details.

\subsection{Evaluations on Balanced Test Set}
\noindent\textbf{Settings}. 
We evaluated the proposed xERM models on balanced test set (\ie~IB=1). All networks were trained for 200 epochs on CIFAR100-LT, 30 epochs on Places365-LT, and 90 epochs on ImageNet-LT and iNaturalist18. 

\noindent\textbf{Metrics}. We used the top-1 accuracy as an overall metric for all the datasets. To further evaluate the unbiasedness, we divided the test set of CIFAR100-LT-IB-100 and ImageNet-LT into three subsets according to the number of samples in each class: {many-shot} (categories with $>$100 images), {medium-shot} (categories with 20$\sim$100 images), and {few-shot} (categories with $<$20 images). 
Previous works focused on the macro-average recall for evaluation, especially on the few-shot subset. 
However, recall alone is not enough to measure the performance on each subset. A trivial solution to achieve a high recall on the few-shot subset is predicting all the samples to the few-shot classes. In this case, the performance on few-shot subset is not reliable. Therefore, we further reported the macro-average precision and the F1 scores for a more comprehensive evaluation. 
The results are summarized in Table~\ref{tab:cifar100IB100}-\ref{tab:inaturalist18}. Further evaluations on Places365-LT and iNaturalist18 are in Appendix.

\input{tables/CIFAR100LT100IB}
\input{tables/ImageNetLT}
\input{tables/CIFARPlacesiNat}

\noindent\textbf{Results on top-1 accuracy}. Overall, xERM outperforms state-of-the-art methods in terms of the top-1 accuracy on all datasets. 
In particular, compared to the balanced models, xERM model consistently outperforms TDE by $2.1\%$, $2.3\%$, $1.1\%$, and $1.4\%$, and PC by $1.4\%$, $4.3\%$, $2.2\%$ and $1.1\%$ on CIRAR100-LT-IB-100, ImageNet-LT, Places365-LT and iNaturalist18.
These results demonstrate the effectiveness and generalizability of our xERM.

\noindent\textbf{Results on recall}. When it comes to the recall performances on the subsets, \textit{all of the previous methods enhance the overall accuracy at the expense of a large head performance drop.}
For example, in Table~\ref{tab:cifar100IB100}, although LADE~\cite{LADE} achieves a high top-1 accuracy of $45.0\%$ and few-short recall of $29.3\%$, its many-shot recall encounters with a significant decrease of $8.4\%$ compared to the baseline XE. 
In contrast, our xERM achieves the highest overall performance with a competitive head performance against XE. 
Surprisingly, xERM$_{\!T\!D\!E}$ is the first to beat XE w.r.t the recall performances on both CIFAR100LT-IB-100 and ImageNet-LT. These results demonstrate that our xERM achieves the unbiaseness from the factual and the counterfactual model, \ie, improving the overall accuracy without scarifying the head recall.

\noindent\textbf{Results on precision and F1 score}. We further reported precision and F1 score to give a more comprehensive evaluation.
According to the results on CIFAR100-LT-IB-100 and ImageNet-LT, the \textit{low precision and high recall} are observed on (1) \textit{many}-shot subset for the imbalanced model (XE); (2) \textit{few}-shot set for the balanced models.
This indicates that the imbalanced models are ``head'' biased and the balanced models are ``tail'' biased.
In contrast, on ImageNet-LT, our xERM$_{\!P\!C}$ achieves both high recall and precision (recall: $67.6\%$, precision: $53.1\%$) on the many-shot subset compared to factual model XE that achieves high recall with lower precision (recall: $68.6\%$, precision: $46.9\%$), and achieves high recall and precision on the few-shot subset  (recall: $24.0\%$, precision: $52.4\%$) compared to PC~\cite{LADE} that achieves competitive recall with lower precision (recall: $23.8\%$, precision: $32.0\%$).
These results demonstrate the unbiasedness of our proposed method.
Besides, an unbiased model should have a balanced precision over the three subsets. For example, in Table~\ref{tab:ImageNetLT}, our xERM$_{\!P\!C}$ manifests $53.1\%$, $55.0\%$ and $52.4\%$ over many-shot, medium-shot and few-shot precision with a small variance, while XE manifests $46.9\%$, $59.1\%$ and $60.7\%$ and PC manifests $56.3\%$, $49.7\%$ and $32.0\%$ with large variances. 
As a result, our xERM achieves the highest F1 score on all the three subsets.

\subsection{Evaluations on Imbalanced Test Set}

\input{tables/ShiftedCifar}
\input{tables/ShiftedImageNet}
\input{tables/feature}

\noindent\textbf{Motivation}. Most state-of-the-art methods assume the test sets are well-balanced (\ie, IB$=$1) and leverage this prior to improve the performance, \ie, playing a ``prior flipping'' game. 
As most long-tailed classification methods lift up the overall accuracy at the expense of head performance loss, evaluation on balanced test set alone cannot reveal their shortcoming.
We considered a new setting where evaluations are conducted under long-tailed test set with different imbalance ratios. This setting is complementary to the balanced test for the unbiasedness evaluation, \ie, performing well on both imbalanced and balanced test distributions. 

In addition, in real-world applications, the test distributions are usually unknown to us, and the classes are often long-tailed distributed because of Zipf's law of nature, \eg, cancer detection. 
Evaluation under long-tailed test set is closer to the wild.
Note that this setting is different from~\citet{LADE}, where the test class prior is known during the training stage. In our setting, no side information about the test distribution prior is accessible to the models.

\noindent\textbf{Settings}. The model architectures are the same as the ones for balanced test and the test splits for imbalanced test are similarly long-tailed like the train set. We established the long-tailed test splits by downsampling the original well-balanced test set with various imbalanced ratios, which is the same as the training set construction.

\noindent\textbf{Results}. Table~\ref{tab:LTtestCIFAR100} and~\ref{tab:LTtestImageNet} show the top-1 accuracy on various long-tailed test splits. As expected, XE performs well when the test set is also imbalanced. Compared to XE, all the previous balanced methods enjoy performance gains on the well-balanced test set while achieves much lower accuracies on the imbalanced test splits. 
For example, in Table~\ref{tab:LTtestCIFAR100}, when IB$=$100, same as the training set, the accuracies of all balanced models are at least $5.6\%$ less than that of XE.

Note that our xERM outperforms all the other balanced methods in every imbalance setting, and achieves competitive results against TE method even when testing on extreme imbalanced ratio, \eg, IB$=$100 on the CIFAR100 test split. When the imbalance ratio of the train set and test set are different, the performance of xERM is even higher than that of XE. 
These results demonstrate the advantage of xERM: \textit{our proposed method is agnostic to both test distributions, \ie, it performs well on balanced and imbalanced test set.} 

\subsection{Evaluations on Feature Representation} \label{sec:feature}

\noindent\textbf{Settings}. We conducted an empirical study on the quality of the feature representations. Specifically, we fixed the backbone for feature extraction, and re-trained the classifier head on a balanced dataset. Take CIFAR100-LT as an example. The models were first trained on the CIFAR100-LT dataset with IB$=$100. Then we fixed the backbone, and fine-tuned the classifiers on the balanced full CIFAR100 train set. 
Table~\ref{tab:feature} compares the performances on three benchmarks and we used ResNet-32, ResNet-50 and ResNeXt-50 as backbones for CIFAR100, Places365 and ImageNet, respectively. 
\input{images/feat_tSNE}
\input{images/cam}

\noindent\textbf{Results on the empirical study}. The classifiers built upon the xERM backbones consistently yield large improvement in terms of accuracy, recall and precision across the three datasets. xERM$_{\!P\!C}$ gains $2.7\%, 2.8\%$ and $3.7\%$ on top-1 accuracy on CIFAR100, Places365 and ImageNet datasets, respectively. These results indicate that the xERM indeed produces better feature representation.
Moreover, as expected, the performance 
gap between XE and TDE backbones is marginal. 
\textit{This indicates that TDE does not improve the quality of feature representation, and it merely adjusts the biased decision boundary to achieve overall improvement. }
Also, PC does not modify the training process, which leads to the same backbone as XE-training.
As comparisons, our xERM achieves consistent improvements on top of both TDE and PC. 

\noindent\textbf{Results on t-SNE visualization.} The t-SNE visualization~\citep{van2008visualizing} of embedding space on CIFAR100-LT test set is plotted in Figure \ref{fig:tSNE}. We selected 9 classes and their class legends are listed from head (top) to tail (bottom). Compared with baseline TDE~\citep{TDE}, feature embedding of xERM$_{\!T\!D\!E}$ is more compact and better separated, \eg, ``television'' and ``wardrobe'' are overlapped in the feature embedding space of TDE (\cf Figure \ref{fig:tSNE} (a)), but they are more separable in the feature embedding space of xERM$_{\!T\!D\!E}$  (\cf Figure \ref{fig:tSNE} (b)).

\noindent\textbf{Results on CAM visualization}. Figure~\ref{fig:cam} further visualized the activation maps using CAM~\citep{cam}. Compared to TDE and PC, our xERM$_{\!T\!D\!E}$ and xERM$_{\!P\!C}$ yield more compact heatmap.
For example, for ``Loafer'', our method pays more attention on shoes rather than the legs; in column ``ski'', our method focuses on the skier instead of the snow background. The visualization indicates that the predictions of xERM are less likely to be influenced by context region.

\subsection{Ablations}\label{sec:exp-ab}
\input{images/ablation}

\noindent \textbf{Analysis of weights.} Recall that the weights $\wf$ and $\wcf$ control the balance between two cross-entropy losses. Some readers may wonder whether the improvement comes from knowledge distillation, which shares similar formulation except for using constant weights $w \in  [0,1]$ and $1-w$ for $\wcf$ and $\wf$ to blend the KL divergence loss and cross-entropy loss. In particular, $w\!=\!0$ denotes that model is only trained using the XE loss while $w\!=\!1$ denotes self-distillation~\cite{zhang2019your}. Figure~\ref{fig:ablation}~(a) illustrates the results of using constant weights ranging from 0 to 1. As shown in Figure~\ref{fig:ablation}~(a), knowledge distillation indeed can improve the performance compared to XE. However, it exhibits a large performance gap with xERM. This indicates that xERM is not a simple knowledge distillation approach.

\noindent \textbf{Effect of the parameter $\gamma$.} Recall that the hyper-parameter $\gamma$ with a large absolute value would push $w$ towards 0 or 1. In particular, $\gamma\!=\!0$ is equivalent to 
a joint learning
with $\wf\!=\!\wcf\!=\!0.5$, and $\gamma\!<\!0$ denotes reversing the ratio in Eq.~\eqref{eq:wfcf}. Figure~\ref{fig:ablation}~(b) shows the influence of the value of gamma $\gamma$ on the CIFAR100-LT-IB-50 dataset. Overall, xERM on top of TDE reaches its optimal accuracy with $\gamma\!=\!2$, and consistently outperforms TDE with $\gamma\!>\!0$. When $\gamma$ is set as a negative value, the model underperforms TDE. This observation empirically verifies the correctness of Eq.~\eqref{eq:wfcf}. 



%% file: tables/dataset.tex
\begin{table}[t]
\caption{The details of the long-tailed training set.}
\vspace{-2mm}
\centering
\scalebox{0.85}{
\begin{tabular}{lccc}
\shline
Dataset       & \#classes & \#samples & IB            \\ \hline
CIFAR100-LT   & 100       & 50K       & \{10, 50, 100\} \\
Places365-LT  & 365       & 62.5K     & 996           \\
ImageNet-LT   & 1K      & 186K      & 256           \\
iNaturalist18 & 8K      & 437K      & 500           \\ \shline
\end{tabular}
}
\label{tab:dataset}
\vspace{-4mm}
\end{table}

%% file: tables/CIFAR100LT100IB.tex
\begin{table}[t]
\setlength\tabcolsep{2pt}
\centering
\caption{
Comparison with state-of-the-arts on CIFAR100-LT-IB-100 with ResNet-32.}
\vspace{-2mm}
\scalebox{0.85}{
\begin{tabular}{l|c|ccc|ccc|ccc}
\shline
\multirow{2}{*}{Method}   & \multirow{2}{*}{Acc} & \multicolumn{3}{c|}{Recall} & \multicolumn{3}{c|}{Precision} & \multicolumn{3}{c}{F1} \\ 
                                  &       & Many  & Med    & Few   & Many  & Med   & Few   & Many  & Med  & Few   \\ 
\hline
XE                            & 40.5     & 68.5     & 39.9     &  8.6      &  38.5    & 50.9      & \tbf{52.2}  & 49.3  & 44.7  & 14.8 \\   
BBN                & 42.3     & 61.5     & 41.7     &  20.2     &  46.6    & 49.7      & 35.1  & 53.0  & 45.3  &  25.6 \\
cRT          & 44.4     & 62.9     & 45.1     &  21.8     &  46.9    & 48.2      & 36.7  & 53.7  & 46.6  & 27.4   \\
LADE              & 45.0     & 60.1     & 43.3     & \tbf{29.3}&\tbf{52.2}& 50.9      & 32.0 & 55.8  &  46.8 &    30.5  \\
\hline
TDE                 & 44.1     & 63.9     & 46.9     &  17.8     &  46.6    & 45.5      & 34.1  & 53.9  & 46.2  & 23.4  \\
PC             & 45.5     & 62.2     & 45.6     &  25.8     &  50.6    & 48.5      & 32.5  & 55.8  & 47.0  & 28.7  \\ 
\hline
\bf{xERM}$_{\!T\!D\!E}$    & {46.2} &\tbf{69.0}&\tbf{48.8}&  16.7     &  46.4    & 49.3      & 45.5    & 55.5   &\tbf{49.0}& 24.4      \\ 
\bf{xERM}$_{\!P\!C}$ &\tbf{46.9}& 65.5     & 45.4     &  26.8     &  51.1    &\tbf{51.5}& 35.9  & \tbf{57.4}  & 48.3 &\tbf{30.7}    \\ 
\shline
\end{tabular}
}
\vspace{-2mm}
\label{tab:cifar100IB100}
\end{table}

%% file: tables/ImageNetLT.tex
\begin{table}[t]
\setlength\tabcolsep{2pt}
\centering
\caption{
Comparison with state-of-the-arts on ImageNet-LT with ResNeXt-50.}
\vspace{-2mm}
\scalebox{0.85}{
\begin{tabular}{l|c|ccc|ccc|ccc}
\shline
\multirow{2}{*}{Method}   & \multirow{2}{*}{Acc} & \multicolumn{3}{c|}{Recall} & \multicolumn{3}{c|}{Precision} & \multicolumn{3}{c}{F1} \\ 
                                   &               & Many           & Med            & Few   & Many   & Med   & Few   & Many  & Med  & Few   \\ \hline
XE                                 & 49.0          & 68.6           & 42.9           & 15.0  &  46.9  & \tbf{59.1}  & \tbf{60.7}  &  55.7  &  49.7  & 24.1     \\   
$\tau$-Norm      & 49.6          & 61.8           & 46.2           & 27.4  &  52.2  & 48.5  & 43.7   & 56.6   & 47.3   & 33.7   \\
LWS                 & 49.9          & 60.2           & 47.2           & 30.3  &  53.0  & 49.1  & 41.3   & 56.4   & 48.1 &   35.0   \\
LADE                    & 51.7          & 62.6           & 49.0           & 30.4  & 55.3   & 50.5   &  41.2    & 58.7  & 49.7     & 34.9    \\
DiVE                    & 53.1          & 64.1           & 50.4           & \textbf{31.5}  &  -     &  -     &   -     &   -    &  -    &  -   \\
DisAlign           & 53.4          & 61.3           & \textbf{52.2}  & 31.4  &  -     &  -     &   -     &   -    &  -    &  -   \\ \hline
PC                    & 48.9          & 60.4           & 46.7           & 23.8  &  56.3  & 49.7  & 32.0   & 58.3   & 48.2   & 27.3      \\
TDE                  & 51.8          & 62.7           & 49.0           & 31.4  & \tbf{57.3}   & 52.3   &  39.5  & 59.9  & 50.6  & 35.0   \\ \hline
\textbf{xERM}$_{\!P\!C}$  & 53.2          & 67.6           & 49.8           & 24.0  &   53.1   & 55.0   & 52.4  & 59.3  & 51.9  & 33.0   \\ 
\textbf{xERM}$_{\!T\!D\!E}$     & \textbf{54.1} & \textbf{68.6}  & 50.0           & 27.5  &   53.5   & 57.3   & 52.0  & \textbf{60.1}  & \textbf{53.4}  & \textbf{36.0}  \\ \shline
\end{tabular}}
\label{tab:ImageNetLT}
\vspace{-4mm}
\end{table}


%% file: tables/CIFARPlacesiNat.tex
\begin{table}[t]
\setlength\tabcolsep{3pt}
\centering
\caption{The top-1 accuracy on CIFAR100-LT-IB-\{50,10\} with ResNet-32, Places365-LT with ResNet-\{50, 152\} and iNaturalist18 with ResNet-50.}
\vspace{-2mm}
\scalebox{0.85}{
\begin{tabular}{l|ll|l|ll|ll}
\shline
\multicolumn{3}{c|}{CIFAR100LT}                                & \multicolumn{3}{c|}{Places365-LT}                       & \multicolumn{2}{c}{iNaturalist18}              \\ \hline
\multirow{2}{*}{Method} & \multicolumn{2}{c|}{IB} & \multirow{2}{*}{Method} & \multicolumn{2}{c|}{Backbone} & \multirow{2}{*}{Method} & \multirow{2}{*}{Acc} \\ 
                        & 50                & 10               &                         & R50     & R152    &                         &                      \\ \hline
XE                      & 46.5              & 58.9    & XE                             & 28.0      & 30.5    & XE                            & 59.3                 \\
LDAM                    & 46.6              & 58.7    & LFME                           & -         & 36.2    & $\tau$-Norm                   & 65.6                 \\
BBN                     & 47.0              & 59.1    & LWS                            & 36.2      & 37.6    & LWS                           & 65.9                 \\
cRT                     & 48.7              & 59.8    & $\tau$-Norm                    & 36.6      & 37.9    & BBN                           & 66.3                 \\
BKD                     & 49.6              & 61.3    & BKD                            & -         & 38.4    & LA                  & 66.4                 \\
LADE                    & 50.5              & 61.7    & LADE                           & 36.5      & 38.8    & LDAE                          & 66.7                 \\
DiVE                    & 51.1              & 62.0    & DisAlign                       & 37.8      & \tbf{39.3}  & BKD                       & 66.8                 \\ \hline
TDE                     & 50.3              & 59.6    & TDE                            & 37.2      & 38.1 & TDE                              & 63.1                 \\
PC                      & 49.7              & 60.9    & PC                             & 36.1      & 38.3  & PC                              & 66.2                 \\ \hline
\tbf{xERM}$_{\!T\!D\!E}$ & \textbf{52.8}           & 62.2   & \tbf{xERM}$_{\!T\!D\!E}$  & \textbf{38.3}  & 39.0 & \tbf{xERM}$_{\!T\!D\!E}$    & 64.5                 \\
\tbf{xERM}$_{\!P\!C}$ & 51.1   & \textbf{62.5}    & \tbf{xERM}$_{\!P\!C}$   & \textbf{38.3}  & \textbf{39.3}   & \tbf{xERM}$_{\!P\!C}$ & 
\textbf{67.3}                 \\ \shline
\end{tabular}
}
\label{tab:inaturalist18}
\vspace{-4mm}
\end{table}

%% file: tables/ShiftedCifar.tex
\begin{table}[t]
\centering
\caption{Top-1 accuracy over all classes on imbalanced CIFAR100-LT test set.}
\vspace{-2mm}
\label{tab:LTtestCIFAR100}
\scalebox{0.8}{
\begin{tabular}{l|cccc}
\shline
Imbalance ratio                 & 100 & 50   & 10   & 5                       \\ \hline
BBN                             & 58.0 & 56.3 & 52.5 & 49.3                  \\
LADE                            & 59.5 & 57.6 & 53.3 & 51.5                   \\
PC                              & 60.0 & 59.2 & 55.3 & 52.9                   \\
cRT                             & 60.8 & 59.5 & 55.0 & 52.0                    \\
TDE                             & 60.8 & 60.0 & 55.7 & 52.9                  \\ \hline
XE                              & \tbf{66.4} & 63.9 & 56.3 & 51.9                  \\ \hline
\textbf{xERM}$_{\!P\!C}$ & 63.3 & 61.5 & 57.0 & 54.6  \\
\textbf{xERM}$_{\!T\!D\!E}$    & \tbf{66.4} & \tbf{64.8} & \tbf{59.3} & \tbf{55.5}     \\ \shline
\end{tabular}}
\end{table}

%% file: tables/ShiftedImageNet.tex
\begin{table}[t]
\centering
\caption{Top-1 accuracy over all classes on imbalanced ImageNet-LT test set.}
\vspace{-2mm}
\label{tab:LTtestImageNet}
\scalebox{0.8}{
\begin{tabular}{l|cccc}
\shline
Imbalanced ratio                  & 50            & 25            & 10            & 5                                             \\ \hline
$\tau$-Norm                       & 59.6          & 58.2          & 56.2          & 54.6         \\
LWS                               & 60.6          & 59.2          & 57.0          & 55.0           \\
PC                                & 58.2          & 56.8          & 54.5          & 52.7           \\
LADE                              & 61.8          & 60.6          & 58.6          & 56.8           \\ 
TDE                               & 63.0          & 61.6          & 59.5          & 57.6           \\ \hline
XE                                & \textbf{67.7} & 65.2          & 61.4          & 58.0           \\ \hline
\textbf{xERM}$_{\!P\!C}$ & 66.8          & 65.3          & 62.5          & 60.1           \\
\textbf{xERM}$_{\!T\!D\!E}$    & \textbf{67.7} & \textbf{66.0} & \textbf{63.5} & \textbf{60.9}   \\ \shline
\end{tabular}}
\vspace{-4mm}
\end{table}

%% file: tables/feature.tex
\begin{table}[t]

\centering
\caption{Studies on the effectiveness of feature representation. }
\vspace{-2mm}
\setlength\tabcolsep{2pt}
\label{tab:feature}
\scalebox{0.85}{
\begin{tabular}{ccccccccccc}
\shline
\multicolumn{1}{l|}{\multirow{2}{*}{Backbone}} & \multicolumn{1}{l|}{\multirow{2}{*}{Acc}} & \multicolumn{3}{c|}{Recall}           & \multicolumn{3}{c|}{Precision}        & \multicolumn{3}{c}{F1} \\
\multicolumn{1}{l|}{}                         & \multicolumn{1}{l|}{}                           & Many & Med & \multicolumn{1}{l|}{Few} & Many & Med & \multicolumn{1}{l|}{Few} & Many   & Med   & Few   \\ \hline
\multicolumn{11}{c}{CIFAR100}                                                                                                                                                                            \\ \hline
\multicolumn{1}{l|}{XE (PC)}               & \multicolumn{1}{c|}{52.6} & 60.3 & 51.9 & \multicolumn{1}{c|}{44.4} & 59.6 & 51.1 & \multicolumn{1}{c|}{44.4} &  60.0 & 51.5  & 44.4  \\
\multicolumn{1}{l|}{TDE }  & \multicolumn{1}{c|}{52.6} & 60.4 & 51.7 & \multicolumn{1}{c|}{44.4} & 59.5 & 51.0 & \multicolumn{1}{c|}{44.5} &  60.0 & 51.4  & 44.5  \\ 
\multicolumn{1}{l|}{LADE} & \multicolumn{1}{c|}{53.9} & 58.7 & 53.8 & \multicolumn{1}{c|}{47.8} & 60.2 & 54.5 & \multicolumn{1}{c|}{47.1} & 59.4  & 54.1   & 47.4  \\  \hline
\multicolumn{1}{l|}{\textbf{xERM}$_{\!T\!D\!E}$} & \multicolumn{1}{c|}{55.1} & 62.8 & 54.5 & \multicolumn{1}{c|}{46.7} & 61.7 & 53.9 & \multicolumn{1}{c|}{48.1} &  62.3 & 54.2  & 47.4  \\
\multicolumn{1}{l|}{\textbf{xERM}$_{\!P\!C}$}    & \multicolumn{1}{c|}{\tbf{55.3}} & \bf 60.9 & \bf 56.0  & \multicolumn{1}{c|}{\bf 48.0} & \bf 63.7 & \bf 54.3 & \multicolumn{1}{c|}{\bf 48.3} & \bf 62.3 &\bf 55.1  &\bf 48.1 \\ \hline
\multicolumn{11}{c}{Places365}                                                                                                                                                                           \\ \hline
 \multicolumn{1}{l|}{XE (PC)}    & \multicolumn{1}{c|}{43.8} & 43.8 &44.0 & \multicolumn{1}{c|}{43.5} & 39.9 & 43.5 & \multicolumn{1}{c|}{49.3} & 41.7 & 43.7 & 46.2  \\
\multicolumn{1}{l|}{TDE}  & \multicolumn{1}{c|}{43.8} & 43.8 & 43.9 & \multicolumn{1}{c|}{43.6} & 39.7 & 43.6 & \multicolumn{1}{c|}{48.7} & 41.6 & 43.8 & 46.0  \\ 
\multicolumn{1}{l|}{LADE}  & \multicolumn{1}{c|}{44.3} & 42.9 & 45.9 & \multicolumn{1}{c|}{43.1} & 43.4 & 45.1 & \multicolumn{1}{c|}{45.7} & 43.1 & 45.5  & 44.4  \\
\hline
\multicolumn{1}{l|}{\textbf{xERM}$_{\!T\!D\!E}$} & \multicolumn{1}{c|}{44.6} & 44.1 & 45.3 & \multicolumn{1}{c|}{44.0} & 40.4 & 44.9 & \multicolumn{1}{c|}{49.5} & 42.1 & 45.1 & 46.6   \\ 
\multicolumn{1}{l|}{\textbf{xERM}$_{\!P\!C}$}    & \multicolumn{1}{c|}{\bf 46.6} & \bf 45.1 & \bf 48.2 & \multicolumn{1}{c|}{\bf 46.0} & \bf 44.2 & \bf 49.0 & \multicolumn{1}{c|}{\bf 53.3} & \bf 44.6 & \bf 48.6 &\bf 49.4      \\
\hline
\multicolumn{11}{c}{ImageNet}                                                                                                                                                                            \\ \hline
\multicolumn{1}{l|}{XE (PC)}  & \multicolumn{1}{c|}{56.5} & 64.5 & 53.8 & \multicolumn{1}{l|}{43.2} & 59.8 & 55.1 & \multicolumn{1}{l|}{50.6} & 62.1 & 54.4 &  46.6   \\
\multicolumn{1}{l|}{TDE}  & \multicolumn{1}{c|}{56.5} & 64.4 & 53.8 & \multicolumn{1}{l|}{43.7} & 60.2 & 55.2 & \multicolumn{1}{l|}{49.8} & 62.2 & 54.5 & 46.6   \\
\multicolumn{1}{l|}{LADE}  & \multicolumn{1}{c|}{57.9} & 62.6 & 55.7 & \multicolumn{1}{l|}{52.2} & 62.4&56.5 & \multicolumn{1}{l|}{52.9} & 62.5 & 56.1 & 52.5 \\
\hline
\multicolumn{1}{l|}{\textbf{xERM}$_{\!T\!D\!E}$} & \multicolumn{1}{c|}{58.9} & 66.5 & 56.4 & \multicolumn{1}{c|}{46.2} & 62.1 & 57.8 & \multicolumn{1}{c|}{\bf 63.2} & 64.2 &  57.1 & 49.4  \\ 
\multicolumn{1}{l|}{\textbf{xERM}$_{\!P\!C}$}    & \multicolumn{1}{c|}{\bf 60.2} &\bf 64.8 & \bf 58.2 & \multicolumn{1}{c|}{\bf 53.8} & \bf 64.9 &\bf 58.3 & \multicolumn{1}{c|}{ 53.9} &\bf 64.8 & \bf 58.2 & \bf 53.8 \\ 
\shline
\end{tabular}}
\vspace{-4mm}
\end{table}

%% file: images/feat_tSNE.tex
\begin{figure}[t]
    \centering
    \includegraphics[width=0.4\textwidth]{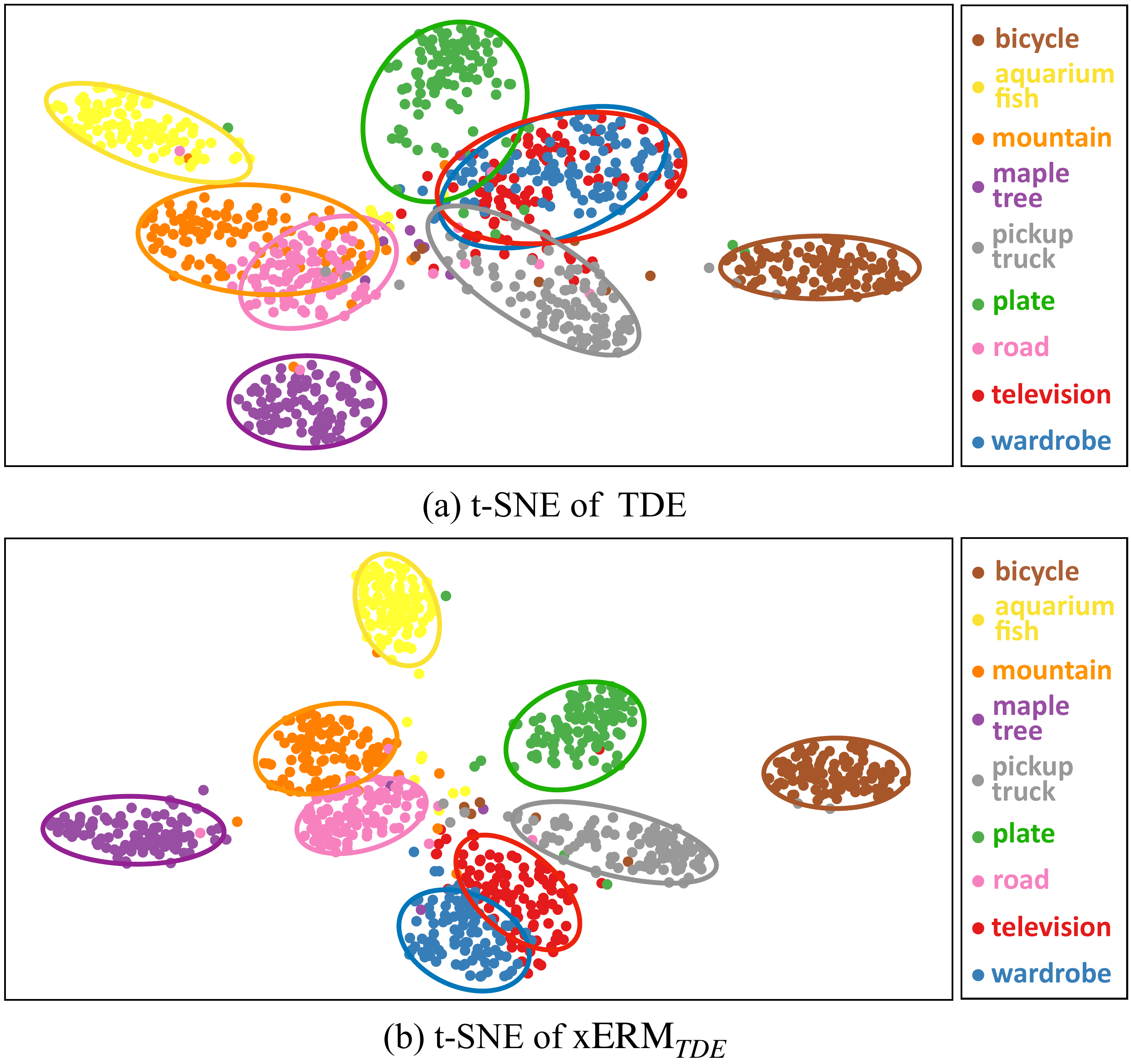}
    \vspace{-2mm}
    \caption{t-SNE visualizations on CIFAR100-LT. (a) visual features learned by TDE and (b) our {xERM}$_{\!T\!D\!E}$.}
    \label{fig:tSNE}
\vspace{-4mm}
\end{figure}

%% file: images/cam.tex
\begin{figure}[t]
    \centering
    \includegraphics[width=0.48\textwidth]{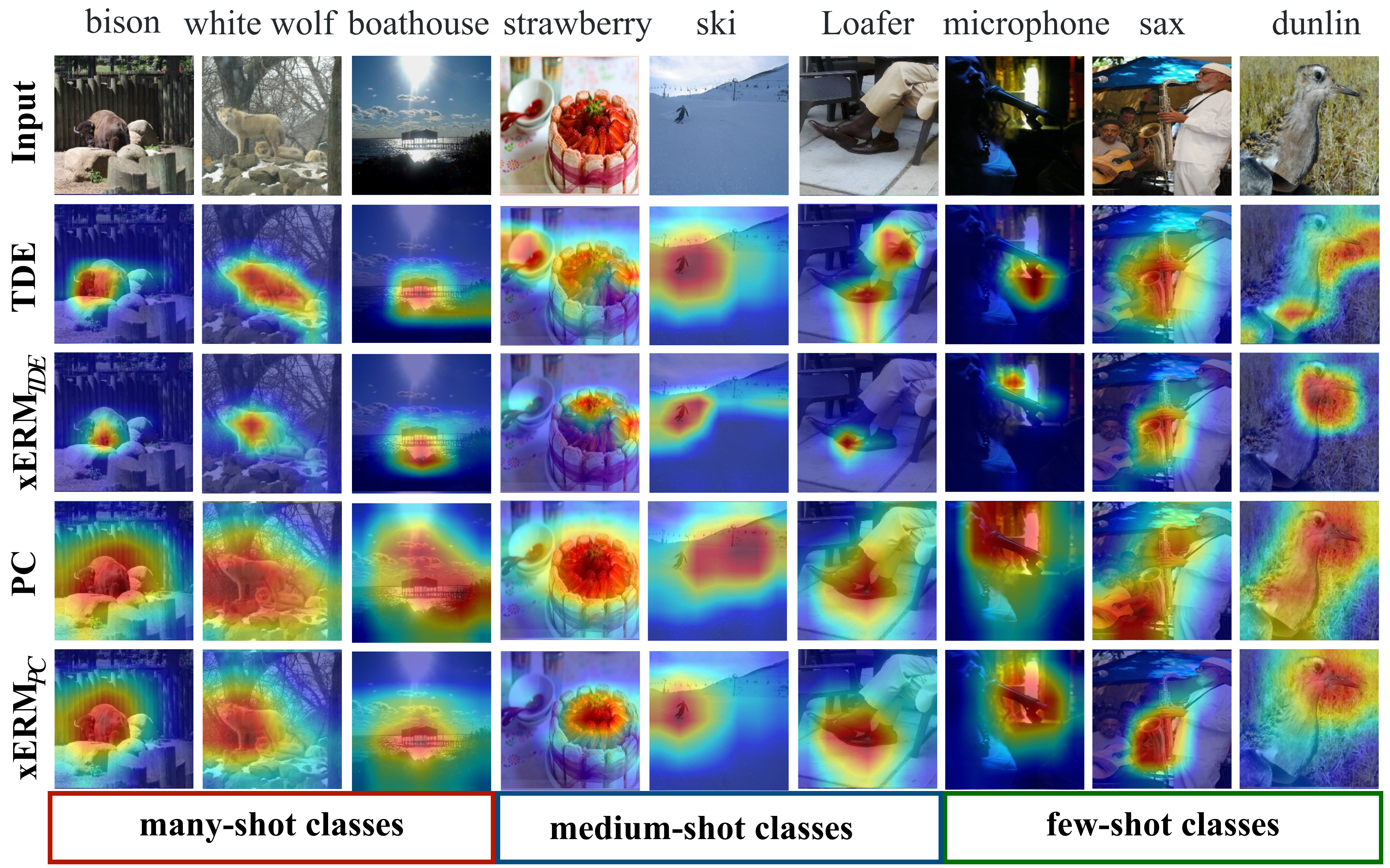}
    \vspace{-4mm}
    \caption{CAM visualizations on ImageNet-LT.}
    \label{fig:cam}
\vspace{-4mm}
\end{figure}

%% file: images/ablation.tex
\begin{figure}[t]
    \centering
    \includegraphics[width=0.48\textwidth]{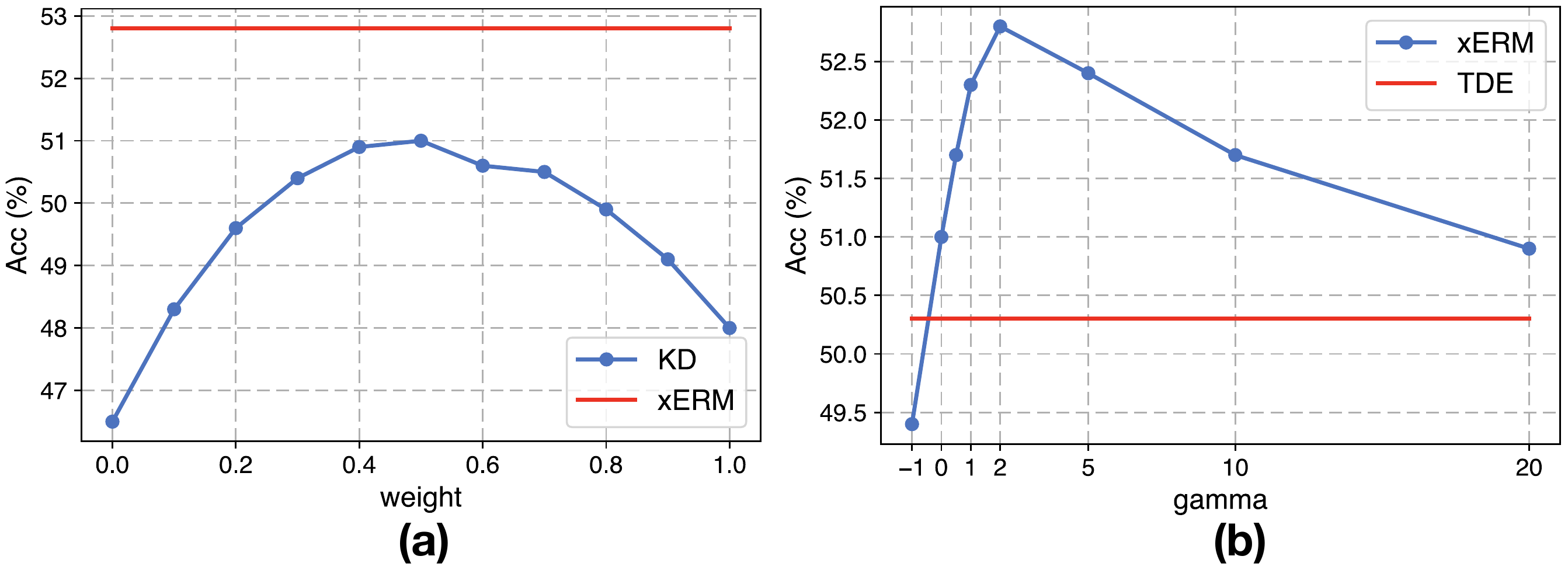}
    \caption{(a) Comparisons with a varying constant weight $w$. on CIFAR100-LT-IB-50 dataset. (b) Comparisons with a varying gamma $\gamma$ in Eq.~\eqref{eq:wfcf} on CIFAR100-LT-IB-50 dataset.}
    \label{fig:ablation}
\vspace{-4mm}
\end{figure}

%% file: sections/6-conclusion.tex
\section{Conclusion}
Despite the overall accuracy improvement on the balanced test set, today's long-tailed models are still far from ``improved'' and exhibit the bias towards tail classes. 
In this work, we proposed xERM framework to train an unbiased test-agnostic model for long-tailed classification. 
Grounded by a causal theory, xERM removes the bias via minimizing the cross-domain empirical risk. 
Extensive experiments demonstrate that the strong improvement on both balanced and imbalanced test sets indeed comes from better feature representation rather than catering to the tail.
The limitations of xERM are three-fold, which will be addressed in future work: 1) it relies on the quality of the balanced model of the unseen balanced domain, 2) the counterfactual is currently only able to imagine a balanced distribution, and 3) our two-stage training pipeline involves two extra models, which increases the computation cost during training. We believe this work provides a new perspective to diagnose the unbiasedness of long-tailed classification and will inspire future works.

%% file: sections/7-ack.tex
\section*{Acknowledgments}
The authors would like to thank the reviewers for their comments that help improve the manuscript. 
This research is supported by the National Research Foundation, Singapore
under its AI Singapore Programme (AISG Award No: AISG2-PhD-2021-01-002).

%% file: sections/a1-proof.tex
\section*{A1 Proof}

We provide a detailed explanation of Eq. \eqref{eq:pygdox} here. The backdoor adjustment of $P(y|do(x))$ \textit{w.r.t.} the confounder $S$ is written as: 

\begin{align}
    &\pygdox\\
    \label{eq:withdo} &=\sum\limits_{S=s\in\{0,1\}}P(y|do(x),S=s)P(S=s|do(x))\\ 
    \label{eq:wodo} &=\sum\limits_{S=s\in\{0,1\}}P(y|x,S=s)P(S=s)\\
    &=\frac{\pxys}{\pxs}\ps+\frac{\pxyns}{\pxns}\pns\\
    &=\frac{\pxys}{\pxgs}+\frac{\pxyns}{\pxgns}
\end{align}

Since the confounder $S$ is under controlled, \ie, $S\!=\!0$ or $S\!=\!1$, we can remove the $do()$ operator in Eq. \eqref{eq:withdo} to obtain Eq. \eqref{eq:wodo} and remove $do(x)$ in $P(S=s|do(x))$.

%% file: sections/a2-implementation.tex
\section*{A2 Experimental Setup}
\input{tables/breakdown}
\input{tables/Places365-ResNet50}
\input{tables/iNaturalist18}

\subsection*{A2.1 Details of Datasets.}

In order to analyze the performance in each subset, we also counted the number of categories per subset in Table~\ref{tab:breakdown}. We can see that the medium-shot and few-shot classes dominate the iNaturalist18 train set, where the trade-off between the ``head'' and ``tail'' contribution is more obvious (\cf Table \ref{tab:iNaturalist}).

\subsection*{A2.2 Evaluation Metrics}
We reported the macro-average recall, precision and F1 score for each subset. We first computed the recall for each class and then calculate the average of each subset. The recall for many-shot subset is denoted as:
\begin{equation}
    \mathrm{Recall}_\mathrm{many}=\frac{1}{|\mathcal{S}_\mathrm{many}|}\sum_{i\in \mathcal{S}_\mathrm{many}}\frac{T\!P_i}{T\!P_i+F\!N_i}
\end{equation}
where $\mathcal{S}_\mathrm{many}$ represents the classes of the many-shot subset, $T\!P_i$ is the number of true positive samples of class $i$ and $F\!N_i$ is the number of false negative samples of class $i$. Similarly, the precision for many-shot subset is denoted as: 
\begin{equation}
    \mathrm{Precision}_\mathrm{many}=\frac{1}{|\mathcal{S}_\mathrm{many}|}\sum_{i\in \mathcal{S}_\mathrm{many}}\frac{T\!P_i}{T\!P_i+F\!P_i}
\end{equation}
where $F\!P_i$ is the number of false positive samples of class $i$. We can further compute the F1 score for many-shot subset as:
\begin{equation}
    \mathrm{F1}_\mathrm{many}=2\cdot \frac{\mathrm{Recall}_\mathrm{many}\cdot \mathrm{Precision}_\mathrm{many}}{\mathrm{Recall}_\mathrm{many}+\mathrm{Precision}_\mathrm{many}}
\end{equation}
The recall, precision and F1 score for medium-shot and few-shot subsets can be computed analogously.

\subsection*{A2.3 Construction of Imbalanced Test Set}
We constructed the imbalanced test set by downsampling the original balanced test set. 
Assume the class labels are sorted by descending the number of train samples per class, the number of samples for each class of the imbalanced test set can be defined as: $n_i=N\cdot \mu^{(i-1)/(C-1)}$, where $C$ is the number of classes, $n_i$ is the number of test samples for class $i \in [1,C]$, $N$ is the number of test samples per class in the original balanced test set and $\mu$ is the reciprocal of the imbalanced ratio. 
The each class of the original CIFAR100 test set has 100 samples, hence the maximum imbalance ratio is 100. 
Analogously, we can construct the long-tailed test set for ImageNet with the maximum ratio of 50 respectively.

\subsection*{A2.4 Implementation Details}
We followed the original configurations of TDE~\cite{TDE} and PC~\cite{LADE} to train the balanced models, and disabled the post-hoc operation to obtain their imbalanced models. The official implementation and licence of TDE and PC can be found in:
\begin{itemize}
  \item \href{https://github.com/KaihuaTang/Long-Tailed-Recognition.pytorch}{\texttt{https://github.com/KaihuaTang/\\Long-Tailed-Recognition.pytorch}}
  \item \href{https://github.com/hyperconnect/LADE}{\texttt{https://github.com/hyperconnect/LADE}} 
\end{itemize}

For our xERM models, we use the vanilla SGD optimizer with momentum $0.9$ for all experiments.

For CIFAR100-LT, we used ResNet-32 with a batch size of 256 on one 1080Ti GPU. The models are trained for 200 epochs with multi-step learning rate scheduler with warm-up like BBN~\cite{BBN}. The learning rate is initialized as 0.2 and decayed by $0.01$ at epoch 120 and 160 respectively.    

For Places365-LT, we used ResNet-\{50,152\} with a batch size of 256 and 128 respectively on four 1080Ti GPUs.
The models are pretrained on the full ImageNet dataset, following~\cite{Decouple}. We trained the models for 30 epochs with multi-step learning rate scheduler, which decays by 0.1 at epoch 10, 20 respectively. The learning rate is initialized as 0.1 for the classifier and 0.01 for the backbone. 

For ImageNet-LT, we used ResNeXt-50~\cite{xie2017aggregated} with a batch size of 256 on four 1080Ti GPUs. The learning rate is decayed by a cosine learning rate scheduler from 0.05 to 0 in 90 epochs. 

For iNatrualist18, we used ResNet-50 with a batch size of 256 on four Tesla-P100 GPUs. The learning rate is decayed by a cosine learning rate scheduler from 0.1 to 0 in 90 epochs.

All the datasets are publicly available for downloading, we provide the download links as follows:
\begin{itemize}
    \item \href{https://www.cs.toronto.edu/~kriz/cifar.html}{\texttt{https://www.cs.toronto.edu\\/$\sim$kriz/cifar.html}}
    \item \href{http://places2.csail.mit.edu/download.html}{\texttt{http://places2.csail.mit.edu\\/download.html}}
    \item \href{https://image-net.org/}{\texttt{https://image-net.org/}}
    \item \href{https://github.com/visipedia/inat_comp/tree/master/2018}{\texttt{https://github.com/\\visipedia/inat\_comp/tree/master/2018}}
\end{itemize}
The implementation code of our xERM is available at  \url{https://github.com/BeierZhu/xERM}, the use of our code is released under BSD-3. 

%% file: tables/breakdown.tex
\begin{table}[h]
\centering
\caption{Number of classes for the three subsets.}
\scalebox{0.85}{

\begin{tabular}{l|ccc|c}
\shline
Dataset            & Many  & Med  & Few & Total \\ \hline
CIFAR100-LT-IB-100 &  35   &  35   & 30   &  100    \\
Places365-LT       &  131  & 163   &  71   &  365     \\
ImageNet-LT        &  385  & 479   & 136   &  1000     \\
iNaturalist18      &  842  & 4076  & 3224  &  8142    \\ \shline
\end{tabular}}
\label{tab:breakdown}
\end{table}

%% file: tables/Places365-ResNet50.tex
\begin{table*}[t]
\centering
\caption{
Comparison with state-of-the-arts on Places365-LT.}
\vspace{1em}
\scalebox{0.85}{
\begin{tabular}{l|c|ccc|ccc|ccc}
\shline
\multirow{2}{*}{Methods}   & \multirow{2}{*}{Acc} & \multicolumn{3}{c|}{Recall} & \multicolumn{3}{c|}{Precision} & \multicolumn{3}{c}{F1} \\ 
                                   &               & Many           & Med            & Few   & Many   & Med   & Few   & Many  & Med  & Few  \\ \hline
XE                                 & 27.6          & \bf 46.7       &  22.9          & 3.2   & 30.3   & \bf 45.4  & \bf 60.5  & 36.7  & 30.4 & 6.1  \\ 
LWS~\cite{Decouple}                & 36.2          & 40.2           &  36.4          & 28.2  & 34.0   &  38.7     & 41.5      & 36.8  & 37.5 & 33.6  \\
$\tau$-Norm~\cite{Decouple}        & 36.6          & 34.4           &  38.4          &\bf 36.4  & 38.1   & 35.9      & 32.4   & 36.2  & 37.1 & 34.3     \\
LADE~\cite{LADE}                   & 36.5          & 42.8           &  38.8          & 19.7  & 36.6   & 36.8      & 34.6      & 39.5  & 37.7 & 25.1   \\ 
DisAlign~\cite{DisAlign}           & 37.8          & 39.3           &  40.7          & 28.5  & -      & -         & -         & -     & -    &  -    \\ \hline
TDE~\cite{TDE}                     & 37.2          & 36.4           &  39.0          & 34.4 & \bf 42.1   & 39.8  & 35.1  & 39.0  & 39.4 & 34.4     \\ 
PC~\cite{LADE}                     & 36.1          & 43.7           &  39.1          & 15.5  & 36.3   & 36.4  & 36.6  & 39.7  & 37.7 & 21.8 \\\hline 
\textbf{xERM}$_{\!T\!D\!E}$     & \bf 38.3      & 41.0           &  38.8          & 32.4  & 38.3   & 40.8  & 39.3  & 39.6  &  39.7 & \bf 35.5  \\ 
\textbf{xERM}$_{\!P\!C}$        & \bf 38.3      & 45.0           &  \bf 41.8      & 18.2  & 36.1   & 39.8  & 50.8  & \bf 40.1  & \bf 40.8 & 26.8  \\ \shline
\end{tabular}}
\vspace{1em}
\label{tab:Places365R50}
\end{table*}

%% file: tables/iNaturalist18.tex
\begin{table*}[t]
\centering
\caption{
Comparison with state-of-the-arts on iNaturalist18.}
\vspace{1em}
\scalebox{0.85}{
\begin{tabular}{l|c|ccc|ccc|ccc}
\shline
\multirow{2}{*}{Methods}   & \multirow{2}{*}{Acc} & \multicolumn{3}{c|}{Recall} & \multicolumn{3}{c|}{Precision} & \multicolumn{3}{c}{F1} \\ 
                              &               & Many           & Med            & Few   & Many   & Med   & Few   & Many  & Med  & Few   \\ \hline
XE                            & 61.6          & \bf 71.4       & 63.2           & 57.1  & 39.9   & 71.2  & \bf 82.9  & 51.2  & 67.0 & 67.6  \\ 
$\tau$-Norm~\cite{Decouple}   & 65.6          & 65.6           & 65.3           & 65.9  & 58.7   & 69.3  & 73.6  & 61.9  & 67.2 & 69.5    \\
LWS~\cite{Decouple}           & 66.0          & 65.5           & 66.5           & 65.5  & 58.9   & 69.5  & 75.1  & 62.0  & 68.0 & 70.0   \\ 
BBN~\cite{BBN}                & 66.4          & 49.4           & \bf 70.8       & 65.3  & 66.6   & 67.6  & 79.1  & 56.7  & 69.2 & 71.5      \\
LADE~\cite{LADE}              & 66.7          & 65.4           & 66.3           & \bf 67.5 & 63.8   & 70.8  & 73.0  & 64.6  & 68.5 & 70.2    \\ \hline
TDE \cite{TDE}                & 63.1          & 68.4           & 65.5           & 58.8  & 61.4   & 68.5  & 71.0  & 64.7  & 67.0 & 64.3  \\ 
PC \cite{LADE}                & 66.2          & 63.7           & 66.2           & 66.9  & \bf 61.7   & 70.1  & 73.3  & 62.7  & 68.1 & 70.0  \\ \hline
\textbf{xERM}$_{\!T\!D\!E}$& 64.5          & 70.4           & 67.5           & 59.4  & 60.7   & 70.4  & 75.6  & \bf 65.2  & 68.9 & 66.5   \\
\textbf{xERM}$_{\!P\!C}$   & \bf 67.3      & 70.3           & 68.4       & 66.1  & 54.2   & \bf 71.8  & 78.1  & 61.2  & \bf 70.1 & \bf 71.6   \\ \shline
\end{tabular}}
\vspace{1em}
\label{tab:iNaturalist}
\end{table*}

%% file: sections/a3-experiments.tex
\section*{A3 Additional Experiments}
\input{tables/appendix_feature}

\subsection*{A3.1 More Results on Balanced Test Set}\label{subsec:balancedTest}
We reported the recall, precision and F1 score for Places365-LT and iNaturalist18. The XE results are obtained by disabling the post-hoc operation of PC~\cite{LADE}.
Table \ref{tab:Places365R50}-\ref{tab:iNaturalist} summarized the detailed evaluations on Places365-LT and iNaturalist18 respectively using ResNet-50 backbone.

\noindent\textbf{Results on accuracy.} Overall, our xERM outperforms state-of-the-art methods including its balanced models in terms of top-1 accuracy, which demonstrates its effectiveness and generalizability. 

\noindent\textbf{Results on recall.} As mentioned before, previous methods enhance the overall accuracy at the expense of a large head performance drop. In Table \ref{tab:Places365R50}, the many-shot recall of $\tau$-Norm and TDE decrease significantly by $12.3\%$ and $10.3\%$ compared to the baseline XE. This phenomenon is particularly obvious in iNaturalist18. According to the last row of Table \ref{tab:breakdown}, the number of the many-shot classes is 842 while the rest 7300 classes belongs to medium-shot and few-shot classes, a large head performance drop can be easily overwhelmed by the improvement on medium-shot and few-shot subset.
For example, in Table \ref{tab:iNaturalist}, although BBN achieves a high top-1 accuracy of $66.4\%$ and the highest medium-shot recall of $70.8\%$, its many-shot recall encounters a dramatic drop from $71.4\%$ (XE) to $49.4\%$. 
 In contrast, our xERM$_{\!P\!C}$ achieves much higher recall ($45\%$ on Places365-LT and $70.4\%$ on iNaturalist18) on many-shot subset compared to other balanced models.

\noindent\textbf{Results on precision and F1 score.} Furthermore, in Table \ref{tab:iNaturalist}, our xERM$_{\!P\!C}$ again achieves competitive recall and high precision (recall: $70.3\%$ precision: $54.2\%$) compared to factual model XE that achieves high recall with lower precision (recall: $71.4\%$ precision: $39.9\%$), and achieves high recall and precision on few-shot subset (recall: $66.1\%$ precision: $78.1\%$) compared to PC~\cite{LADE} that achieves similar recall with lower precision (recall: $66.9\%$ precision: $73.3\%$). As a result, our xERM achieves the highest F1 score on the two datasets.

\subsection*{A3.2 More Analysis on the Effectiveness of Feature Representation}
We further conducted an empirical study to re-confirm that the strong performance comes from better feature representation rather than the ``head'' and ``tail'' trade-off. Concretely, we trained our xERM on imbalanced datasets under two settings: (1) only train the classifier and use the backbone of XE to extract features; (2) train both the backbone and the classifier. We trained the xERM models on CIFAR100-LT-IB-100 dataset accordingly, and reported the results in Table \ref{tab:appendix_feature}.

xERM models that only retrain the classifier manifest marginal differences against their balanced models. This indicates that xERM cannot improve the overall performance by just adjusting the classifier, it indeed enhance the quality of feature representation.

\subsection*{A3.3 Error Bars on CIFAR100-LT Dataset}

\input{tables/error_bars}

We reported the standard deviation (error bars) on CIFAR100-LT-IB-\{10,50,100\} datasets.
In our experiments, the randomness mainly comes from three three aspects: the construction of train sets, the training process of the imbalanced/balanced models, and the training process of our xERM models. To better evaluate the randomness of our xERM models, we fixed the train sets, imbalanced models and balanced models and ran the experiment for xERM models for 4 times. The results are shown in Table \ref{tab:error_bar}.

\subsection*{A3.4 Results of ensemble models.} \citet{wang2020long} proposed a multi-experts model named RIDE for long-tailed classification. Our xERM framework can be easily applied to RIDE to validate its effectiveness on ensemble models. Note that the implementation of re-balanced method is open in RIDE, thus we applied LDAM loss~\cite{LDAM} and PC to RIDE as our baseline, which are denoted as RIDE-LDAM-$m$ and RIDE-PC-$m$ ($m$ is the number of experts). The training settings follow \citet{wang2020long}. Our learned xERM models share the same model architecture with their balanced models, which are denoted as xERM$_{\text{RIDE-LDAM-}m}$ or xERM$_{\text{RIDE-PC-}m}$. Table \ref{tab:ensemble} shows the results on balanced test set of CIFAR100-LT-IB-100, where our xERM models surpass their corresponding baselines in terms of accuracy, achieving new state-of-the-arts. Compared with baseline models, our xERM improves many-shot/medium-shot recall, medium-shot/few-shot precision and achieves competitive performance on few-shot recall and many-shot precision.

\input{tables/ensemble}

%% file: tables/appendix_feature.tex
\begin{table*}[t]
\centering
\caption{
Studies on the effectiveness of feature representation on CIFAR100-LT-IB-100.}
\vspace{1em}
\scalebox{0.83}{
\begin{tabular}{l|cc|c|ccc|ccc|ccc}
\shline
\multirow{2}{*}{Methods}   & \multicolumn{2}{c|}{Retraining?} & \multirow{2}{*}{Acc} & \multicolumn{3}{c|}{Recall} & \multicolumn{3}{c|}{Precision} & \multicolumn{3}{c}{F1} \\ 
                          & Backbone & CLS  &       & Many  & Med    & Few   & Many  & Med   & Few   & Many  & Med  & Few   \\ 
\hline
XE                   &    &             & 40.5     & 68.5     & 39.9     &  8.6      &  38.5    & 50.9      & \tbf{52.2}  & 49.3  & 44.7  & 14.8 \\   
\hline
TDE\cite{TDE}     &    &           & 44.1     & 63.9     & 46.9     &  17.8     &  46.6    & 45.5      & 34.1  & 53.9  & 46.2  & 23.4  \\
PC \cite{LADE}    &    &            & 45.5     & 62.2     & 45.6     &  25.8     &  50.6    & 48.5      & 32.5  & 55.8  & 47.0  & 28.7  \\ 
\hline
\bf{xERM}$_{\!T\!D\!E}$    &    &  \cmark    & 44.0   & 64.1     & 46.9 &  17.1      & 45.7    &45.5      &35.0     & 53.4  & 46.2 & 23.0  \\ 
\bf{xERM}$_{\!P\!C}$      &    & \cmark      & 45.9   &    60.8  & 46.3 & \bf 28.0   & \bf 52.5    & 48.6     & 32.7    & 56.3  & 47.3 & 30.2    \\ \hline
\bf{xERM}$_{\!T\!D\!E}$     & \cmark   &  \cmark   & {46.2} &\tbf{69.0}&\tbf{48.8}&  16.7     &  46.4    & 49.3      & 45.5    & 55.5   &\tbf{49.0}& 24.4      \\ 
\bf{xERM}$_{\!P\!C}$    & \cmark   &  \cmark &\tbf{46.9}& 65.5     & 45.4     &  26.8     &   51.1    &\tbf{ 51.5}& 35.9  & \tbf{57.4}  & 48.3 &\tbf{30.7}    \\ 
\shline
\end{tabular}
}
\label{tab:appendix_feature}
\end{table*}

%% file: tables/error_bars.tex
\begin{table}[h]
\centering
\caption{Mean and standard deviation on CIFRA100-LT.}
\begin{tabular}{l|lll}
\shline
Imbalance Ratio  & 100 & 50 & 10 \\ \hline
\textbf{xERM}$_{\!T\!D\!E}$ & 46.2\tiny{$\pm$ 0.191} & 52.8\tiny{$\pm$ 0.112} & 62.2\tiny{$\pm$ 0.141}   \\
\textbf{xERM}$_{\!P\!C}$  & 46.9\tiny{$\pm$ 0.150} & 51.1\tiny{$\pm$ 0.096}  & 62.5\tiny{$\pm$ 0.222}    \\ \shline
\end{tabular}
\vspace{1em}
\label{tab:error_bar}
\end{table}

%% file: tables/ensemble.tex
\begin{table*}[h]
\centering
\caption{
Result of ensemble model RIDE on CIFAR100-LT-IB-100.}
\scalebox{0.85}{
\begin{tabular}{l|c|ccc|ccc|ccc}
\shline
\multirow{2}{*}{Methods}   & \multirow{2}{*}{Acc} & \multicolumn{3}{c|}{Recall} & \multicolumn{3}{c|}{Precision} & \multicolumn{3}{c}{F1} \\ 
                                  &       & Many  & Med    & Few   & Many  & Med   & Few   & Many  & Med  & Few   \\ 
\hline
RIDE-LDAM-2  &  47.0   & 66.4  & 48.8   & 22.2  & 51.5  & 47.0  & 34.7 & 58.0 & 47.9 & 27.1 \\  
RIDE-LDAM-3  &  48.3   & 68.0  & 49.7   & 23.5  & 53.0  & 49.0  & 36.5 & 59.6 & 49.3 & 28.6  \\
RIDE-PC-2    &  46.1   & 62.6  & 45.9   & 27.2  & 58.1  & 50.0  & 26.8 & 60.3 & 47.9 & 27.0  \\
RIDE-PC-3    &  47.8   & 64.0  & 46.6   & \textbf{30.1}  & \textbf{59.9} & {51.9} & 28.2 & \textbf{61.9} & 49.1   & 29.1   \\
\hline
\textbf{xERM}$_{\text{RIDE-LDAM-}2}$  & 48.1 &  68.8 & 50.4 & 21.3 & 50.7 & 48.8 & 37.7 & 58.4 & 49.6 &  27.2     \\ 
\textbf{xERM}$_{\text{RIDE-LDAM-}3}$  & 49.6 &  \textbf{70.0} & 51.3 & 23.7 & 52.8 & 49.7 & 41.0 & 60.2 & 50.5 & 30.0  \\ 
\textbf{xERM}$_{\text{RIDE-PC-}2}$    & 48.9 &  64.9 & 51.3 & 27.3 & 56.3 & 52.3 & \textbf{34.0} & 60.3 & 51.8 & 30.3     \\ 
\textbf{xERM}$_{\text{RIDE-PC-}3}$    & \textbf{50.1} &  66.3 & \textbf{51.8} & 29.3 & 57.1 & \textbf{53.4} & 33.8 & 61.4 & \textbf{52.6} & \textbf{31.4} \\ 
\shline
\end{tabular}
}
\label{tab:ensemble}
\end{table*}